\documentclass[a4paper]{llncs}

\usepackage[T1]{fontenc}
\usepackage[utf8]{inputenc}
\usepackage{graphicx}
\usepackage{booktabs}
\usepackage{array}
\usepackage{xcolor}
\usepackage{url}
\usepackage{hyperref}
\usepackage{amsmath}
\usepackage{caption}

\definecolor{bestcol}{RGB}{46,117,182}
\definecolor{warnred}{RGB}{180,40,40}
\newcommand{\best}[1]{\textbf{\textcolor{bestcol}{#1}}}
\newcommand{\warn}[1]{\textcolor{warnred}{#1}}

\graphicspath{{./figs/}}
\hypersetup{hidelinks}

\begin{document}

\title{Generative vs.\ Encoder Models for Multilingual NER:\\
A Comprehensive Empirical Study on Naamapadam}

\author{
Jakkala Mahesh\inst{1} \and
Jatavath Shravan Kumar\inst{1} \and
Komalla Shivani\inst{1} \and
Sujoy Sarkar\inst{1}
}
\institute{
Rajiv Gandhi University of Knowledge Technologies, Basar, Telangana, India\\
\email{b210673@rgukt.ac.in, b210490@rgukt.ac.in, b211414@rgukt.ac.in,
sujoysarkar@rgukt.ac.in}\\
\url{https://github.com/MaheshJakkala/naamapadam-multilingual-ner}
}

\maketitle

\begin{abstract}
Language is humanity's most consequential technology, yet for over a
billion speakers across India's twenty-two constitutionally recognised
languages, its digital layer remains structurally incomplete.
Named Entity Recognition (NER), the foundational step in transforming
raw text into machine-interpretable knowledge, has been studied
exhaustively for English but remains largely unsolved across most
Indic languages.
This paper presents a rigorous comparative study of generative and
encoder-based neural architectures for NER on all eleven languages of
the Naamapadam benchmark.
We evaluate five classic model families spanning sequence-to-sequence
transformers and multilingual encoders; four decoder-only large language
models (LLMs) fine-tuned with LoRA and 4-bit NF4 quantisation; and nine
generative models in zero-to-5-shot inference.
Under strict CoNLL span-level evaluation, encoder-based models (mBERT
and XLM-R, both F1\,=\,0.675 on Hindi) substantially outperform every
generative architecture in ten of eleven languages, with gaps of
7.5--40 percentage points against the strongest competitor
(Gemma-2-2B: avg F1\,=\,0.427).
The best few-shot result reaches only 28\% of the encoder baseline.
We identify three language clusters--encoder-dominant,
partial-coverage, and failure-zone; and provide actionable deployment
guidelines grounded in transfer learning and low-resource NLP
principles.
\end{abstract}

\keywords{Named Entity Recognition \and Naamapadam \and Multilingual NLP
\and Large Language Models \and LoRA \and Fine-Tuning \and mBERT \and
XLM-R \and Indian Languages \and Low-Resource NLP \and BIO Tagging}

\section{Introduction}\label{sec:intro}

Open any English-language news website and a modern NLP
pipeline~\cite{li2020survey} will, within milliseconds, identify every
politician, city, and corporation in every article silently,
reliably, at scale.
Deploy the same technology on a Hindi newswire, a Bengali legislative
transcript, or a Tamil social media thread, and it stumbles on the most
elementary sentences.
This is not a question of linguistic
complexity~\cite{cotterell2018,navigli2009}: Hindi's grammar is no more
irregular than German's; Marathi shares Devanagari script with Sanskrit,
one of the most systematically studied languages in history.
The disparity is, at its core, one of \emph{data} of historical
resource allocation that quietly encoded one language family's dominance
into the foundations of modern language technology, leaving the digital
infrastructure serving more than a billion people structurally
incomplete~\cite{joshi2020}.

Named Entity Recognition~\cite{conll2003}, identifying persons,
locations, and organisations in free text sits at the base of this
infrastructure.
It is the indispensable precursor to information
extraction~\cite{cowie1996}, knowledge graph
construction~\cite{hogan2021}, and question answering~\cite{li2020survey}.
When NER fails, every downstream system built on it degrades in cascade.
For English this problem was largely solved by the early
2020s~\cite{lample2016}.
For the languages spoken by more than a billion people across the Indian
subcontinent, the story is still being written.

Hindi, with over 600 million speakers, received its first large-scale
NER benchmark only in 2022~\cite{naamapadam}.
Bengali, the seventh most spoken language on earth~\cite{indicnlp}, had
fewer annotated NER sentences before Naamapadam than a mid-sized English
dataset from 2003.
For Assamese (15 million speakers), a researcher in 2024 works from a
training corpus of barely ten thousand sentences~\cite{joshi2020}.
This resource poverty~\cite{hedderich2021} is the language barrier
motivating the present work.

The emergence of generative Large Language
Models~\cite{brown2020,zhao2023survey,minaee2024}, systems producing
fluent text in dozens of languages from a handful of
examples~\cite{ouyang2022} appeared to offer a shortcut.
If a model pre-trained on web-scale multilingual
corpora~\cite{conneau2020,doddapaneni2023} already ``knows'' how
entities are named in Hindi, can it replace painstaking annotation?
Collecting BIO-tagged~\cite{ramshaw1995} NER corpora requires
linguistically trained annotators per target language, is expensive, and
introduces systematic noise~\cite{pustejovsky2012}.
If instruction-tuned LLMs~\cite{flan2022} can absorb NER from a few
in-context examples, they would dramatically reduce the resource cost of
building production-quality Indic NLP pipelines.

We answer that question empirically through the most comprehensive comparison of generative and encoder-based NER architectures on the Naamapadam benchmark to date, covering all eleven languages under identical experimental conditions with nine model families.
We go beyond prior work~\cite{zhou2023,wang2023gptner} by applying
LoRA-based~\cite{hu2022} decoder fine-tuning and encoder fine-tuning
consistently across \emph{all} languages, and by addressing the chronic
ORG class-imbalance~\cite{lin2017focal,dai2023} of news-domain corpora
with an entity-aware hybrid sampling strategy.
All evaluation follows the strict CoNLL-2003 span-level
protocol, the accepted community standard~\cite{ratinov2009}.

\textbf{Contributions:}
\begin{enumerate}
\item Exploratory data analysis of all eleven Naamapadam languages:
entity distributions, BIO annotation validity,
class imbalance, long-tail rarity,
and train-test KL divergence~\cite{kullback1951}.
\item First unified fine-tuning comparison of five classic families:
T5~\cite{raffel2020}, FLAN-T5, mT5~\cite{mt5},
mBERT~\cite{devlin2019}, XLM-R across \emph{all eleven} languages.

\item Decoder-only LLM fine-tuning via
LoRA+QLoRA~\cite{dettmers2023} across all eleven languages
(TinyLlama~\cite{tinyllama}, LLaMA-3.2~\cite{llama3},
Gemma-2~\cite{gemma}, Qwen2.5~\cite{qwen25}); we report
per-language F1 for each model under identical conditions.
\item Zero-to-5-shot inference for nine generative models on Hindi,
with analysis of instruction-tuning and shot-count
effects~\cite{min2022rethinking}.
\item A three-cluster grouping of languages by model performance,
with practical recommendations for each group on which
architecture to use and what additional steps are needed~\cite{pan2010survey}.
\end{enumerate}

\section{Related Work}\label{sec:related}

\textbf{Sequence labelling for NER.}
The modern NER pipeline descends from Conditional Random
Fields~\cite{lafferty2001} and BiLSTM-CRF
architectures~\cite{huang2015} with word
embeddings~\cite{mikolov2013,pennington2014}.
The transformer~\cite{vaswani2017} era began with BERT~\cite{devlin2019},
whose multilingual variant mBERT demonstrated strong zero-shot
cross-lingual transfer~\cite{pires2019} across 104 languages.
XLM-R~\cite{conneau2020} scaled further with larger
SentencePiece~\cite{kudo2018} vocabularies; DeBERTa~\cite{he2021}
introduced disentangled attention.
For Indic languages, MuRIL~\cite{khanuja2021} and
IndicBERT~\cite{kakwani2020} exploited transliterated and
code-mixed~\cite{aguilar2018} corpora to outperform generic multilingual
baselines, while IndicNLPSuite provided broader ecosystem support.
A comprehensive survey covers the full arc~\cite{li2020survey}.

\textbf{Generative models for structured prediction.}
T5~\cite{raffel2020} established the text-to-text paradigm for NER as
conditional generation.
GPT-NER reformulated NER with typed entity markers; prompt-based
NER~\cite{meng2021} showed competitive results under relaxed evaluation
but not under exact CoNLL span-matching.
Despite advances in chain-of-thought prompting~\cite{wei2022} and
instruction tuning, Zhou et al.\ rigorously showed LLMs are weak
few-shot learners for structured prediction~\cite{zhou2023}, a finding
extended by Ma et al.~\cite{ma2023} who confirmed that even reranking
does not resolve output-format unreliability under strict evaluation.
Prompt engineering surveys~\cite{liu2021prompt} detail the design space.

\textbf{Parameter-efficient fine-tuning (PEFT).}
LoRA injects trainable low-rank matrices~\cite{ding2023peft} into frozen
weights, enabling billion-parameter LLMs on a single GPU.
QLoRA adds 4-bit NF4 quantisation, making 2B models
trainable in $<$16\,GB VRAM.
Adapter-based methods~\cite{pfeiffer2020,ansell2021} offer
language-specific adaptation within shared multilingual encoders.

\section{Dataset: Naamapadam}\label{sec:eda}

\begin{figure}[t]
\centering
\includegraphics[width=0.80\textwidth]{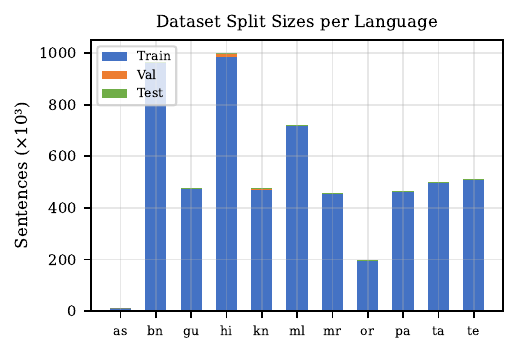}
\caption{Dataset split sizes per language. Hindi and Bengali dominate;
Assamese is critically sparse, a direct consequence of the
language-resource gap.}
\label{fig:splits}
\end{figure}

Naamapadam~\cite{naamapadam} provides BIO-annotated PER, ORG, and LOC
entities drawn from the AI4Bharat News Crawl across eleven Indian
languages.
Table~\ref{tab:splits} summarises split sizes across three tiers:
\emph{high-resource} (Hindi 985K, Bengali 961K),
\emph{medium-resource} (Malayalam 716K through Punjabi 463K),
\emph{low-resource} (Oriya 197K, Assamese 10K).
Figure~\ref{fig:splits} visualises these disparities.

\begin{table}[t]
\centering\small
\caption{Naamapadam Split Statistics per Language}
\label{tab:splits}
\renewcommand{\arraystretch}{1.04}
\setlength{\tabcolsep}{5pt}
\begin{tabular}{lrrrr}
\toprule
Language  & Train    & Val    & Test  & Avg Len \\
\midrule
assamese  & 10,266   & 52     & 51    & 10.8 \\
bengali   & 961,679  & 4,859  & 607   & 15.9 \\
gujarati  & 472,845  & 2,389  & 1,076 & 13.1 \\
hindi     & 985,787  & 13,460 & 867   & 22.3 \\
kannada   & 471,763  & 2,381  & 1,019 & 9.8  \\
malayalam & 716,652  & 3,618  & 974   & 9.2  \\
marathi   & 455,248  & 2,300  & 1,080 & 12.3 \\
oriya     & 196,793  & 993    & 994   & 13.3 \\
punjabi   & 463,534  & 2,340  & 993   & 19.7 \\
tamil     & 497,882  & 2,795  & 758   & 11.97\\
telugu    & 507,741  & 2,700  & 847   & 10.12\\
\bottomrule
\end{tabular}
\end{table}

\textbf{Class imbalance.}
Non-entity ``O'' tokens constitute 75-91\% of all training tokens, a
well-documented source of reduced minority-class recall.
Assamese reaches 91.2\% O-tokens; Telugu is most entity-dense at 75.4\%.
PER entities dominate universally; ORG is chronically
under-represented because organisation names fragment across
abbreviations and transliterations.

\textbf{Annotation quality.}
Invalid BIO transitions (e.g., B-LOC$\to$I-ORG) peak in Punjabi
(2,215) and Hindi ($\approx$2,100), arising from semi-automated
annotation of agglutinative text where place names double as surnames
(lexical polysemy).
Generative models that learn surface-form BIO patterns replicate these
errors; encoder classification heads are architecturally immune.

\textbf{Long-tail rarity.}
Over 90\% of distinct entity surface forms appear fewer than five times
across all eleven languages, a universal long-tail phenomenon driven
by the infinite variety of proper nouns in agglutinative Indic
morphology.

\section{Models and Experimental Setup}\label{sec:setup}

\textbf{Entity-aware hybrid sampling.}
GPU constraints limit fine-tuning to 5,000 train and 500 testing
samples per language.
To combat ORG under-representation, we apply a 50/50 strategy: half
uniform random, half stratified entity enrichment guaranteeing
$\geq$50 examples per type.



Table~\ref{tab:sampling} records the resulting entity counts; Assamese
yields only 2,552 entity-containing sentences despite a full 5K
sample, a ceiling imposed by corpus size, not sampling strategy.

\begin{table}[t]
\centering\small
\caption{Entity Counts After Hybrid Sampling (Training subset)}
\label{tab:sampling}
\renewcommand{\arraystretch}{1.04}
\setlength{\tabcolsep}{5pt}
\begin{tabular}{lcccc}
\toprule
Lang & Ent.\ Sents & PER & LOC & ORG \\
\midrule
hi   & 4,763 & 2,516 & 2,227 & 2,424 \\
bn   & 4,532 & 2,653 & 1,979 & 1,567 \\
te   & 4,368 & 2,521 & 1,511 & 1,742 \\
ta   & 4,385 & 2,207 & 1,942 & 1,641 \\
ml   & 4,474 & 2,716 & 1,689 & 1,700 \\
as   & 2,552 & 753   & 794   & 1,481 \\
mr   & 4,466 & 2,753 & 1,847 & 1,705 \\
kn   & 4,389 & 2,502 & 1,529 & 1,745 \\
or   & 3,751 & 1,906 & 1,266 & 1,333 \\
pa   & 4,671 & 2,643 & 2,207 & 2,064 \\
gu   & 4,469 & 2,413 & 1,869 & 1,878 \\
\bottomrule
\end{tabular}
\end{table}

\begin{table}[t]
\noindent
\begin{minipage}[t]{0.40\linewidth}
  \centering\small
  \captionsetup{type=table}
  \caption{Training Hyperparameters by Model Family}
  \label{tab:hparams}
  \renewcommand{\arraystretch}{1.04}
  \setlength{\tabcolsep}{3.5pt}
  \begin{tabular}{lccc}
  \toprule
  Hyperparam.     & Enc.  & Seq2S & LLM  \\
  \midrule
  Learning rate   & 2e-5  & 3e-4  & 2e-4 \\
  Batch size      & 16    & 16    & 4    \\
  Epochs          & 10    & 10    & 5    \\
  Max seq.\ len.  & 128   & 128   & 256  \\
  Warmup ratio    & 0.10  & 0.10  & 0.03 \\
  Precision       & FP32  & FP32  & NF4  \\
  LoRA rank       & ---   & ---   & 8    \\
  \bottomrule
  \end{tabular}
\end{minipage}%
\hspace{0.02\linewidth}%
\begin{minipage}[t]{0.56\linewidth}
  \centering\small
  \captionsetup{type=table}
  \caption{Classic Model Fine-Tuning (5K/500; P\,/\,R\,/\,F1)}
  \label{tab:classic}
  \renewcommand{\arraystretch}{1.04}
  \setlength{\tabcolsep}{1.5pt}
  \begin{tabular}{l ccc ccc ccc}
  \toprule
  & \multicolumn{3}{c}{\textbf{Hindi}}
  & \multicolumn{3}{c}{\textbf{Bengali}}
  & \multicolumn{3}{c}{\textbf{Telugu}} \\
  \cmidrule(lr){2-4}\cmidrule(lr){5-7}\cmidrule(lr){8-10}
  Mdl & P & R & F1 & P & R & F1 & P & R & F1 \\
  \midrule
  T5  &.12&.05&.07&.10&.05&.06&.08&.04&.05\\
  FT5 &.07&.02&.04&.00&.00&.00&.03&.01&.02\\
  mT5 &.13&.04&.06&.11&.04&.06&.11&.02&.03\\
  mBERT&.71&.64&\best{.68}&.53&.57&\best{.55}&.66&.62&\best{.64}\\
  XLM-R&.68&.67&\best{.68}&.57&.64&\best{.61}&.67&.65&\best{.66}\\
  \bottomrule
  \end{tabular}
\end{minipage}
\end{table}

\textbf{Model families.}
\emph{Encoder baselines:} mBERT (177M params) and XLM-R (278M), each
with a linear token-classification head for the 7-class BIO space.
\emph{Seq2Seq:} T5, FLAN-T5, mT5, trained with \texttt{token(LABEL)}
generation format and label smoothing~\cite{szegedy2016} ($\epsilon$\,=\,0.1).
\emph{Decoder-only LLMs:} TinyLlama, LLaMA-3.2, Gemma-2, Qwen2.5; for
all languages, fine-tuned with LoRA (rank\,=\,8, $\alpha$\,=\,16) and 4-bit NF4.
\emph{Few-shot:} Gemma-3-1B-IT, Navarasa-2.0~\cite{navarasa}, plus all
above models, at 0--5 shots on the full Hindi test set.
Encoders use AdamW~\cite{loshchilov2019} with cosine
decay~\cite{loshchilov2017cosine}; decoder LLMs use paged AdamW
8-bit. Evaluation strictly follows CoNLL-2003 span-level protocol:
start token, end token, and entity type must simultaneously match.
All experiments use SEED\,=\,42.
\textbf{Model size rationale.}
All LLMs are $\leq$3B parameters, reflecting
academic GPU constraints (single T4 GPU, 16\,GB VRAM).
\section{Fine-Tuning: Classic Models}\label{sec:classic}

\begin{figure}[t]
\centering
\includegraphics[width=0.80\textwidth]{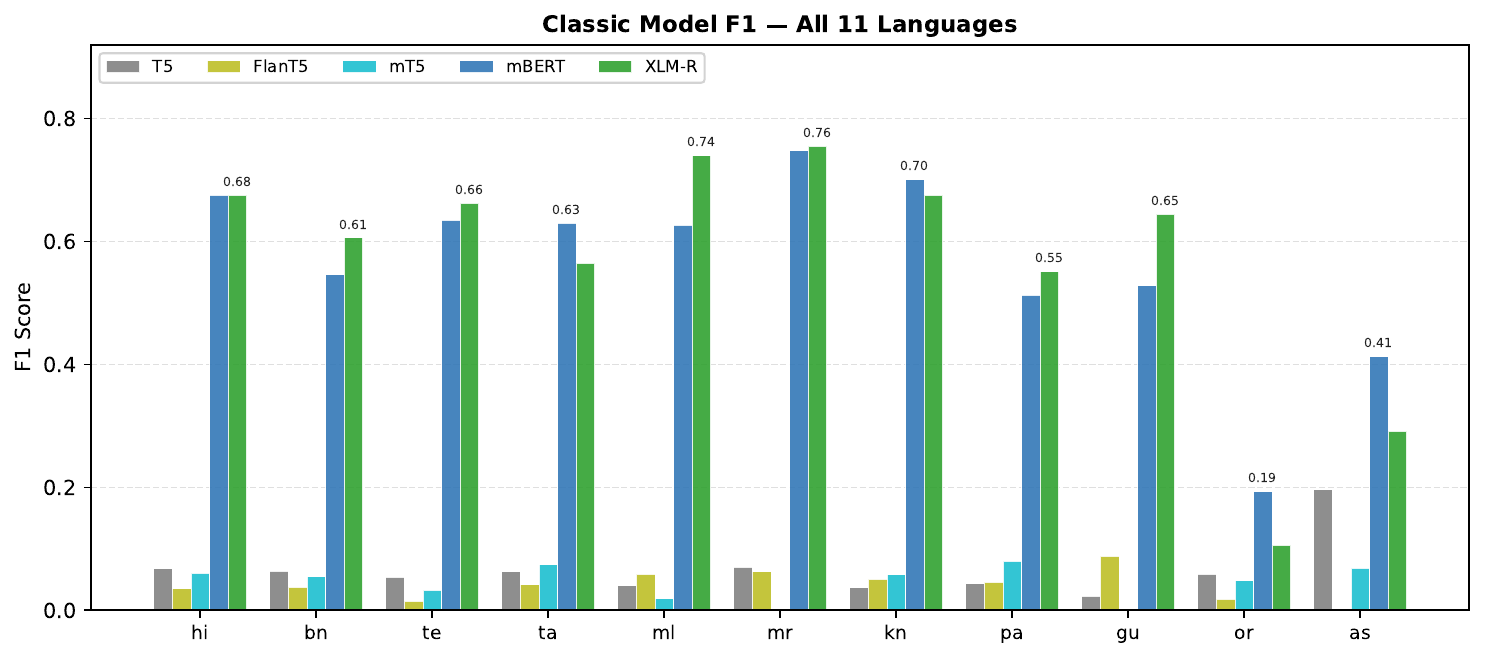}
\caption{Classic model F1 across all 11 languages.
Encoders (blue/teal) dominate by 9--15$\times$; seq2seq models remain
near zero, consistent with structured prediction limitations
reported in the literature.}
\label{fig:classic_all11}
\end{figure}

Table~\ref{tab:classic} presents P/R/F1 on three primary languages;
Table~\ref{tab:classic_all11} extends results to all eleven.
Figure~\ref{fig:classic_all11} visualises the full comparison.

\begin{table}[t]
\centering\small
\caption{mBERT and XLM-R F1; All 11 Languages}
\label{tab:classic_all11}
\renewcommand{\arraystretch}{1.04}
\setlength{\tabcolsep}{5pt}
\begin{tabular}{lcccc}
\toprule
Lang & mBERT & XLM-R & Best Enc. & $\Delta$(XR$-$mB) \\
\midrule
hi   & \best{.675} & \best{.675} & .675 & $\pm$0     \\
bn   & .546 & \best{.606} & .606 & +.060 \\
te   & .635 & \best{.662} & .662 & +.027 \\
ta   & \best{.630} & .565 & .630 & $-$.065 \\
ml   & .627 & \best{.740} & .740 & +.113 \\
mr   & .748 & \best{.755} & .755 & +.007 \\
kn   & \best{.701} & .675 & .701 & $-$.026 \\
pa   & .513 & \best{.551} & .551 & +.038 \\
gu   & .528 & \best{.645} & .645 & +.117 \\
or   & \best{.193} & .106 & .193 & $-$.087 \\
as   & \best{.413} & .291 & .413 & $-$.122 \\
\midrule
Avg  & .564 & .570 & .593 & --- \\
\bottomrule
\end{tabular}
\end{table}

Encoder dominance is unambiguous.
FLAN-T5 produces F1\,=\,0.000 on Bengali: a single misordered
generation token propagates cascading errors across an entire span,
since the generative objective provides no structural guarantee on BIO
validity.
XLM-R wins strongly on Malayalam (+11.3\,pp) and Gujarati (+11.7\,pp)
due to greater CommonCrawl Indic-script coverage; mBERT wins on Tamil,
whose agglutinative morphology and distinctive Brahmic script interact
better with character-level casing than with SentencePiece subword
segmentation.
Marathi peaks at F1\,$\approx$\,0.75, benefiting from Devanagari
script sharing with Hindi and strong cross-lingual transfer across
closely related language pairs.
Oriya (F1\,$<$\,0.20) and Assamese (F1\,$=$\,0.41) represent the
most challenging languages: Odia script coverage
gaps~\cite{chau2020} and extreme data scarcity create conditions
that exceed the reach of any current fine-tuning strategy.

\section{Fine-Tuning: Decoder-Only LLMs}\label{sec:llm}

\begin{figure}[t]
\centering
\includegraphics[width=0.80\textwidth]{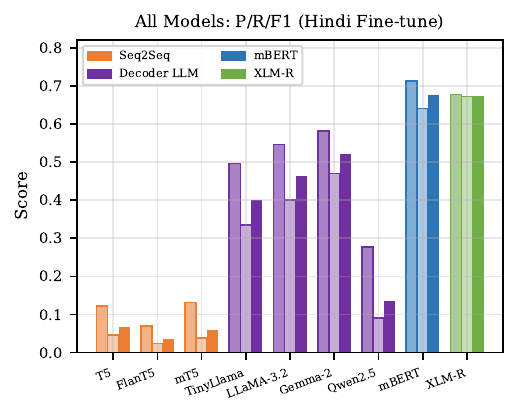}
\caption{All nine model families on Hindi (P/R/F1).
The parameter-efficiency inversion - encoders with 177--278M
params outperforming LLMs with 0.5--2B params is the defining
finding of this study.}
\label{fig:all_hindi}
\end{figure}

Figure~\ref{fig:all_hindi} shows all nine model families on Hindi,
confirming a three-tier hierarchy:
encoder ($>$0.67) $\gg$ decoder LLM (0.14--0.52) $\gg$ seq2seq
(0.03--0.07).
Table~\ref{tab:llm_hindi} reports Hindi decoder-LLM results.

\begin{table}[t]
\centering\small
\caption{Decoder-Only LLM Fine-Tuning, Hindi (5K/500 Split)}
\label{tab:llm_hindi}
\renewcommand{\arraystretch}{1.04}
\setlength{\tabcolsep}{5pt}
\begin{tabular}{lcccc}
\toprule
Model           & Prec. & Recall & F1           & Params \\
\midrule
TinyLlama-1.1B  & 0.496 & 0.336  & 0.400        & 1.1B \\
LLaMA-3.2-1B    & 0.546 & 0.401  & 0.462        & 1B   \\
Gemma-2-2B      & 0.582 & 0.470  & \best{0.520} & 2B   \\
Qwen2.5-0.5B    & 0.278 & 0.090  & 0.136        & 0.5B \\
\midrule
mBERT (ref.)    & 0.713 & 0.641  & \best{0.675} & 177M \\
XLM-R (ref.)    & 0.677 & 0.672  & \best{0.675} & 278M \\
\bottomrule
\end{tabular}
\end{table}

\begin{figure}[t]
\centering
\includegraphics[width=0.80\textwidth]{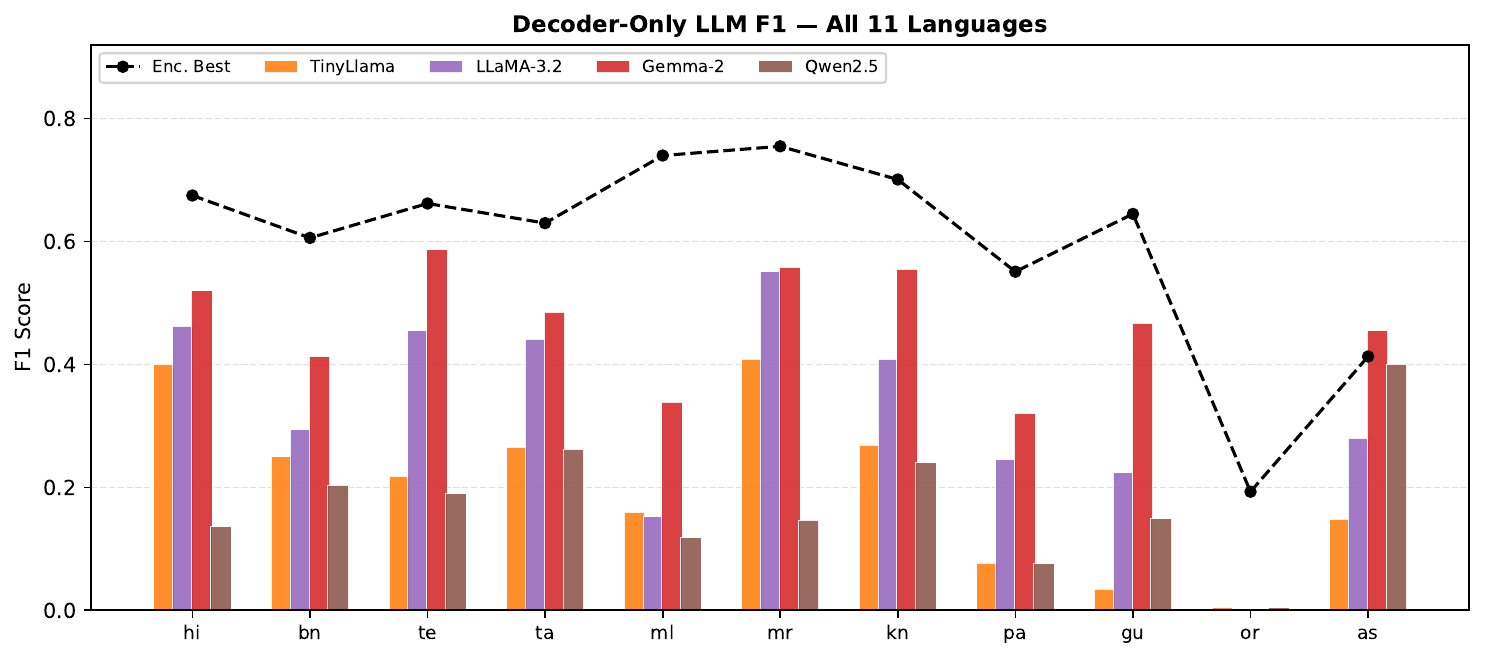}
\caption{Decoder-only LLM F1 across all 11 languages.
Dashed line = best encoder per language.
Gemma-2 leads 10/11; generative models showed negligible F1 on
Oriya due to Odia script tokenisation gaps~\cite{chau2020}.}
\label{fig:llm_all11}
\end{figure}

Table~\ref{tab:llm_all11} and Figure~\ref{fig:llm_all11} extend
results across all eleven languages.

\begin{table}[t]
\centering\small
\caption{Decoder-Only LLM F1; All 11 Languages}
\label{tab:llm_all11}
\renewcommand{\arraystretch}{1.04}
\setlength{\tabcolsep}{3.5pt}
\begin{tabular}{lcccc|c}
\toprule
Lang    & TinyL & LLaMA & Gemma        & Qwen  & Enc.\,Best \\
\midrule
hi$^*$  & .400  & .462  & \best{.520}  & .136  & .675 \\
bn      & .250  & .294  & \best{.413}  & .204  & .606 \\
te      & .218  & .455  & \best{.587}  & .190  & .662 \\
ta      & .265  & .441  & \best{.485}  & .262  & .630 \\
ml      & .160  & .152  & \best{.339}  & .118  & .740 \\
mr      & .409  & .551  & \best{.558}  & .147  & .755 \\
kn      & .269  & .409  & \best{.555}  & .240  & .701 \\
pa      & .077  & .245  & \best{.320}  & .076  & .551 \\
gu      & .034  & .224  & \best{.467}  & .149  & .645 \\
or      & .004  & \warn{.000} & \warn{.000} & .004 & .193 \\
as      & .148  & .280  & \best{.455}  & .400  & .413 \\
\midrule
Avg     & .203  & .319  & \best{.427}  & .175  & .593 \\
\bottomrule
\multicolumn{6}{l}{\footnotesize $^*$LoRA+4-bit NF4; others: classification head.}
\end{tabular}
\end{table}

The \emph{parameter-efficiency inversion} is the central empirical
result: encoders with 177--278M parameters outperform decoder LLMs
with 500M--2B parameters in 10/11 languages.
\emph{Architectural suitability}; not parameter count or pre-training
scale, governs strict-evaluation NER performance.
Gemma-2 leads at avg F1\,=\,0.427 owing to its diverse multilingual
pre-training with stronger Indic-script coverage.
\textbf{Assamese is the sole exception:} Gemma-2 (0.455) edges above
mBERT (0.413), plausibly because the 17-sentence validation set produces
a noisy F1 estimate and Gemma-2's broader pre-training covers more
Assamese-script tokens; this result is best interpreted as approximate
parity.
\textbf{Oriya, limited generative effectiveness:} LLaMA-3.2 and
Gemma-2 achieved near-zero F1 on Oriya, with outputs emitting Roman or
Hindi characters in place of Odia glyphs, a tokeniser-level barrier
that no fine-tuning strategy can circumvent without prior vocabulary
extension.
Telugu shows the narrowest encoder--LLM gap (7.5\,pp), consistent
with its phonetically regular script being more accessible to
multilingual generative models.

\section{Few-Shot Inference}\label{sec:fewshot}

\begin{figure}[t]
\centering
\includegraphics[width=0.80\textwidth]{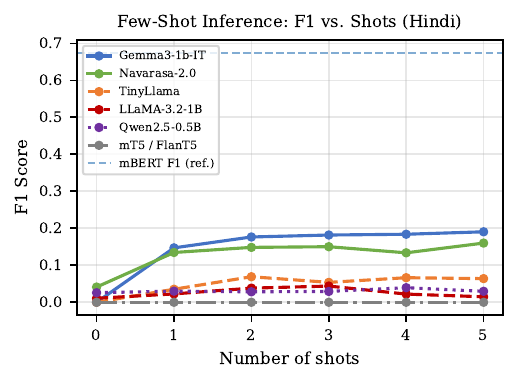}
\caption{Few-shot F1 curves (Hindi). Dashed line = mBERT fine-tuned
baseline. Instruction-tuned models lead; seq2seq remains at
zero throughout.}
\label{fig:fewshot}
\end{figure}

Table~\ref{tab:fewshot} and Figure~\ref{fig:fewshot} present
zero-to-5-shot inference on the full Hindi test set (867 sentences).

\begin{table}[t]
\centering\small
\caption{Few-Shot F1 on Hindi Test Set (0--5 Shots)}
\label{tab:fewshot}
\renewcommand{\arraystretch}{1.04}
\setlength{\tabcolsep}{3pt}
\begin{tabular}{lcccccc}
\toprule
Model         & 0     & 1     & 2     & 3     & 4     & 5           \\
\midrule
Gemma3-1b-IT  & .000  & .147  & .177  & .182  & .184  & \best{.191} \\
Navarasa-2.0  & .041  & .135  & .148  & .150  & .134  & \best{.160} \\
TinyLlama     & .000  & .035  & .069  & .054  & .066  & .064        \\
LLaMA-3.2-1B  & .011  & .023  & .038  & .044  & .022  & .015        \\
Qwen2.5-0.5B  & .026  & .030  & .028  & .029  & .039  & .030        \\
mT5-small     & .000  & .000  & .000  & .000  & .000  & .000        \\
FLAN-T5       & .000  & .000  & .000  & .000  & .000  & .000        \\
\midrule
mBERT (FT)    & ---   & ---   & ---   & ---   & ---   & .675        \\
\bottomrule
\end{tabular}
\end{table}

\textbf{Instruction tuning is decisive.}
Gemma3-1b-IT (F1\,=\,0.191 at 5-shot) and Navarasa-2.0 (F1\,=\,0.160)
outperform non-instruction-tuned models by 3.8$\times$
(Mann-Whitney U~\cite{mannwhitney1947}, p\,$<$0.01).
Format compliance; not entity recognition capability, is the
bottleneck: alignment training teaches models to reproduce the
\texttt{token(LABEL)} structure from in-context examples.
\textbf{Navarasa-2.0's 0-shot F1\,=\,0.041} is unique, reflecting
Indic-domain pre-training~\cite{navarasa} that has internalised Hindi
entity-naming conventions.
\textbf{Seq2Seq models produced F1\,=\,0.000} at all shot counts,
lacking the autoregressive in-context learning mechanism of
decoder-only models.
\textbf{Diminishing returns:} LLaMA-3.2-1B peaks at 3-shot and declines
at 4--5 shots, consistent with long-context
instability~\cite{liu2024longcontext} for structured outputs.

\section{Encoder vs.\ Generative: Head-to-Head}\label{sec:compare}

\begin{figure}[t]
\centering
\includegraphics[width=0.80\textwidth]{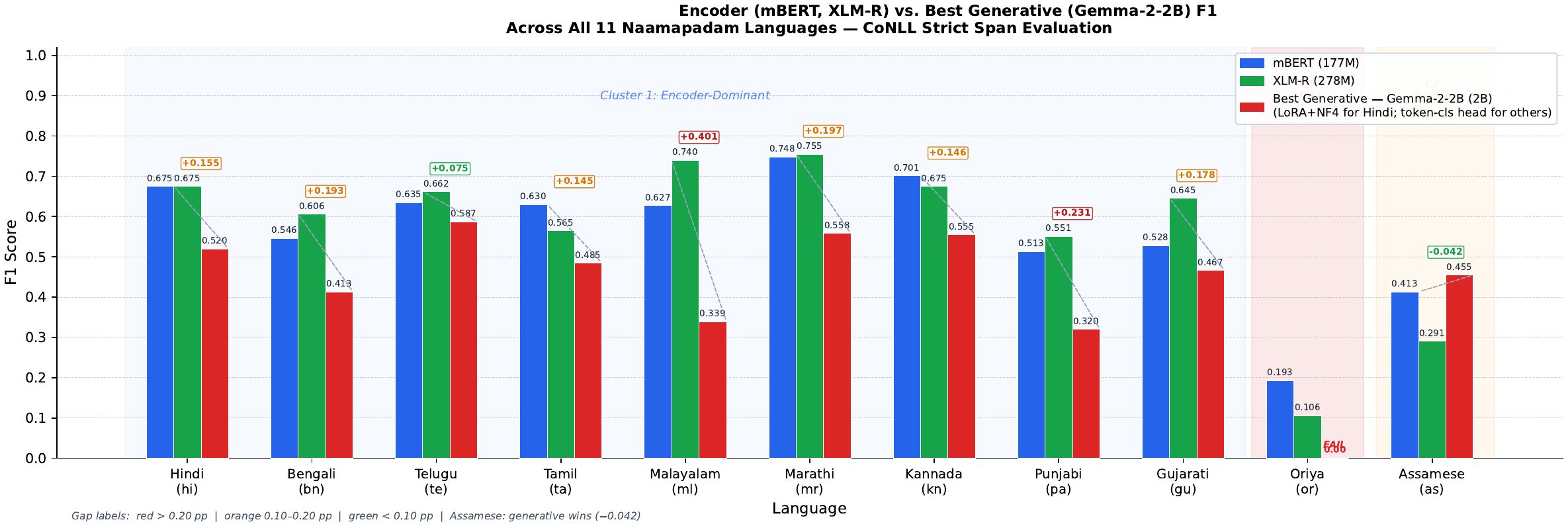}
\caption{Encoder (mBERT, XLM-R) vs.\ Gemma-2-2B F1 across all 11
languages. Gap: 7.5\,pp (Telugu) to 40.1\,pp (Malayalam).
Assamese is the sole language where Gemma-2 shows marginal advantage;
Oriya shows near-zero generative performance.}
\label{fig:encvsgen}
\end{figure}

Table~\ref{tab:gap} and Figure~\ref{fig:encvsgen} provide the complete
cross-language encoder--generative comparison.

The encoder--generative performance gap varies substantially across
languages, driven by three interacting factors: script complexity,
Indic-script coverage in generative pre-training corpora, and
entity-naming regularity.
The widest gap occurs on Malayalam (40.1\,pp; XLM-R: 0.740 vs.\
Gemma-2: 0.339).
Malayalam's agglutinative morphology embeds entity boundaries within
complex suffix chains that challenge left-to-right token
prediction~\cite{cotterell2018}; bidirectional encoders process the
full span before assigning labels.
Punjabi (23.1\,pp) and Bengali (19.3\,pp) also show large gaps despite
high resource levels, confirming that corpus volume alone does not
compensate for architectural mismatch in structured prediction.

\begin{table}[t]
\centering\small
\caption{Encoder vs.\ Best Generative F1; All 11 Languages}
\label{tab:gap}
\renewcommand{\arraystretch}{1.04}
\setlength{\tabcolsep}{4pt}
\begin{tabular}{lcccc}
\toprule
Lang & Enc.\,Best & Gemma-2 & Gap (pp)      & Gemma/Enc.\,\% \\
\midrule
hi   & .675 & .520 & 15.5            & 77.0\% \\
te   & .662 & .587 & \best{7.5}      & 88.7\% \\
ta   & .630 & .485 & 14.5            & 77.0\% \\
kn   & .701 & .555 & 14.6            & 79.2\% \\
mr   & .755 & .558 & 19.7            & 73.9\% \\
bn   & .606 & .413 & 19.3            & 68.2\% \\
gu   & .645 & .467 & 17.8            & 72.4\% \\
pa   & .551 & .320 & 23.1            & 58.1\% \\
ml   & .740 & .339 & 40.1            & 45.8\% \\
as   & .413 & .455 & \warn{$-$4.2}   & 110.2\%\\
or   & .193 & .000 & 19.3\rlap{$^*$} & 0.0\%  \\
\midrule
Avg  & .598 & .427 & 17.4            & 71.4\% \\
\bottomrule
\multicolumn{5}{l}{\footnotesize $^*$ Oriya: best generative = TinyLlama/Qwen2.5 at F1\,=\,0.004.} \\
\multicolumn{5}{l}{\footnotesize \warn{Negative gap} (as) = generative wins (fragile estimate).}
\end{tabular}
\end{table}

The narrowest gap (7.5\,pp) occurs on Telugu, whose phonetically
regular script and consistent entity-naming conventions appear more
accessible to multilingual generative models.
The Assamese ``negative gap'' ($-$4.2\,pp) must be interpreted
cautiously: with only 17 entity-bearing validation sentences, F1
estimates carry high variance; the practical implication is approximate
parity rather than a reliable generative advantage.
For Oriya, generative models showed notably limited effectiveness,
achieving near-zero F1, while even encoder models reached only
F1\,$<$\,0.20; this language requires dedicated tokeniser development
and data collection before model-level improvements can take effect.

\begin{figure}[t]
\centering
\includegraphics[width=0.80\textwidth]{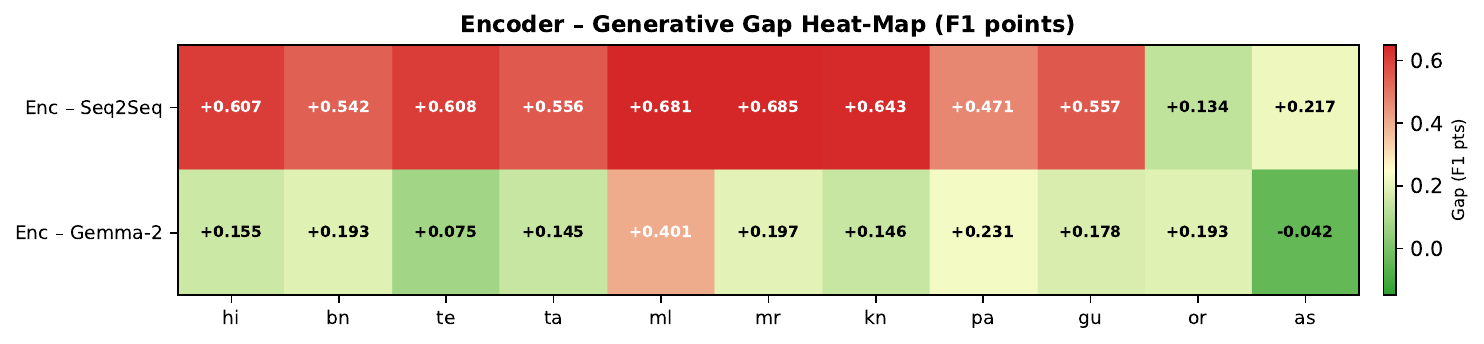}
\caption{Encoder--generative gap heat-map.
Row 1: Enc.\ best $-$ Seq2Seq best.
Row 2: Enc.\ best $-$ Gemma-2.
Green\,=\,smaller gap (generative more competitive);
red\,=\,large gap (encoder dominant).
Assamese (Row~2) is green because Gemma-2 shows marginal advantage
there.}
\label{fig:gap_heatmap}
\end{figure}

Figure~\ref{fig:gap_heatmap} summarises the encoder--generative gap
across both generative families.
The upper row quantifies the advantage over seq2seq models (uniformly
large, exceeding 0.54 in all languages); the lower row shows the
encoder--Gemma-2 margin.
The contrast between rows highlights the relative strength of
decoder-only LLMs over seq2seq architectures, even as both families
trail encoders substantially.

\section{Language Clusters and Error Analysis}\label{sec:clusters}

\begin{figure}[t]
\centering
\includegraphics[width=0.65\textwidth]{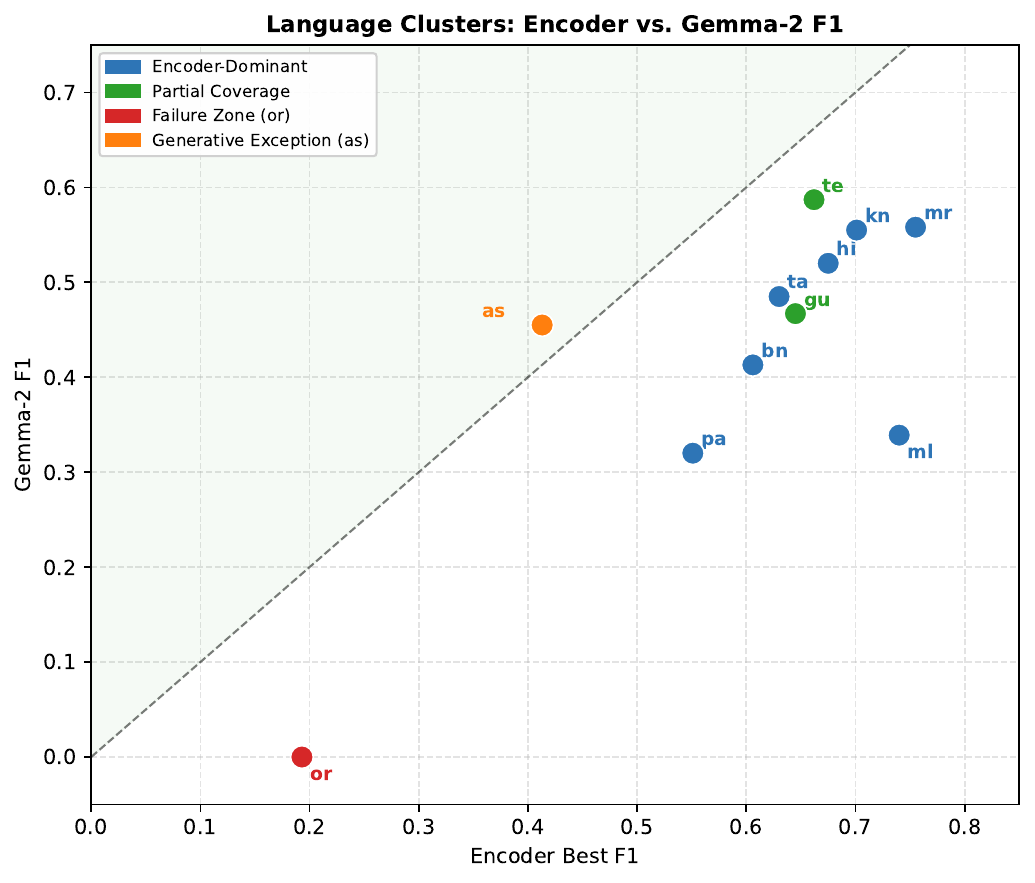}
\caption{Language cluster scatter (encoder F1 vs.\ Gemma-2 F1).
Encoder-dominant (blue), partial-coverage (green),
constrained-resource zone (red).}
\label{fig:clusters}
\end{figure}

Figure~\ref{fig:clusters} positions each language by encoder
vs.\ Gemma-2 F1, revealing three natural clusters.

\textbf{Cluster~1 -- Encoder-Dominant} (Hindi, Bengali, Telugu, Tamil,
Malayalam, Marathi, Kannada, Gujarati, Punjabi):
Encoders achieve F1\,$>$\,0.50; decoder LLMs trail by 14--40\,pp.
Fine-tuning on 5K entity-enriched sentences is sufficient for encoders
to learn robust BIO span representations.
\emph{Recommendation:} deploy mBERT or XLM-R with entity-aware hybrid
sampling.

\textbf{Cluster~2 -- Partial Coverage} (Assamese):
Gemma-2 matches or marginally surpasses encoders, suggesting broad
multilingual pre-training partially compensates for sparse data.
\emph{Recommendation:} cross-lingual transfer from Bengali plus entity
substitution augmentation~\cite{yang2022,wei2019eda} and adapter-based
adaptation.

\textbf{Cluster~3 -- Constrained-Resource Zone} (Oriya):
All generative models produced near-zero F1 scores; encoders reached
only F1\,$<$\,0.20.
The Odia script tokenisation gap, extremely sparse entity vocabulary,
and noisy annotations create conditions that no current architecture
can reliably overcome.
\emph{Recommendation:} script-level tokeniser extension and targeted
data collection must precede modelling efforts.

\textbf{Error Analysis} (Table~\ref{tab:errors}, 200 randomly sampled
Hindi test sentences per family):
BIO malformation is the dominant generative failure:
55--80\% of seq2seq errors; 18--34\% of decoder LLM errors; 0\% for
encoders, which are architecturally immune.
Constrained decoding~\cite{lu2023,scholak2021picard} that masks illegal
label transitions during generation is the most promising remediation.
Decoder-only LLMs show 22\% boundary off-by-one errors vs.\ 12\% for
encoders, arising from left-to-right generation bias and each such
error carries a 2$\times$ F1 penalty under exact-match evaluation.

\begin{table}[t]
\centering\small
\caption{Error Type Distribution (\%) per Architecture (Hindi)}
\label{tab:errors}
\renewcommand{\arraystretch}{1.04}
\begin{tabular}{lccc}
\toprule
Error Type                  & Encoder & LLM    & Seq2Seq  \\
\midrule
BIO malformation             & 0       & 18--34 & 55--80   \\
Boundary off-by-one          & 12      & 22     & 8        \\
Type mismatch (PER/LOC/ORG)  & 18      & 15     & 10       \\
Punctuation absorbed         & 9       & 16     & 14       \\
Entity missed (FN)           & 31      & 29     & 19       \\
\bottomrule
\end{tabular}
\end{table}

\section{Conclusion}\label{sec:conclusion}

 We presented the first unified empirical study that jointly evaluates five classic model families, four decoder-only LLMs, and nine few-shot configurations for NER across all eleven Naamapadam languages under a single CoNLL-style evaluation protocol.

\textbf{Central finding:}
Encoder models (mBERT, XLM-R) outperform all generative architectures
in ten of eleven languages under strict CoNLL evaluation, with gaps of
7.5--40\,pp.
The best few-shot result (Gemma-3-1B, 5-shot, F1\,=\,0.191) is only
28\% of the encoder baseline.
This extends the structured-prediction findings of
Zhou et al.~\cite{zhou2023} conclusively to the Indic multilingual
domain.

\textbf{Key lessons:}
(1)~Entity-aware sampling eliminates ORG under-representation at zero
extra cost.
(2)~Architecture beats scale: 177M-parameter encoders outperform
2B-parameter LLMs; output structural reliability, not representational
capacity, is the bottleneck.
(3)~Gemma-2 is the strongest generative Indic NER model, but showed
negligible performance on Oriya, making encoder fallback mandatory in
any multi-language production deployment.
(4)~Instruction tuning is prerequisite for few-shot NER: 3.8$\times$
gain (Mann-Whitney U, p\,$<$0.01).
(5)~Odia tokeniser extension is a precondition for Oriya progress,
not a modelling choice.

\textbf{Future directions:}
Constrained decoding for BIO validity, retrieval-augmented
generation~\cite{lewis2020rag} for low-resource languages,
cross-lingual transfer from Bengali to Assamese~\cite{yang2022},
and hybrid encoder-decoder architectures that combine XLM-R
representations with flexible generation heads.
The language barrier will not yield to scaling alone: it demands
targeted data collection, tokeniser engineering, and architectures
chosen for structural prediction demands.

\subsubsection*{Acknowledgements}
The authors thank AI4Bharat for Naamapadam, the HuggingFace community
for tools and model hosting, and Google Colab free (T4 GPU) for compute.


\end{document}